\documentclass{article}
\usepackage{graphicx}
\begin{document}
\title{{On the Smallness of the Large Language Models Scaling Exponents}}
\author{Sauro Succi$^{1,2}$, Peter V.\ Coveney$^3$ and Alex Hansen$^2$\\ \\
$^1$Italian Institute of Technology,\\ 
Viale Regina Elena, 291, 00161, Rome, Italy\\  \\
$^2$PoreLab, Physics Department,\\ 
Norwegian University of Science and Technology,\\
7491 Trondheim, Norway\\ \\
$^3$ Centre for Computational Science, Chemistry Department,\\
University College of London,\\
20 Gordon Street, WC1H 0AJ, London, United Kingdom
}
\maketitle

\begin{abstract}
We discuss reasons why the scaling exponents of current Large 
Language Models (LLMs) applications are indicating an unsustainable 
regime in terms of energy resources.
We further show that attributing the smallness of such exponents
to a numerical bias due to the neglect of a non-zero value
of the loss function in the limit of infinite data
(``pedestal effect") does not remove the unsustainability issue.
Finally, the effects of the smoothness (roughness) of the data
on the scaling exponents is commented upon based on an analogy
with phenomenological models of fluid turbulence.
\end{abstract}

\section{Introduction}

AI and most notably transformers-based large Language Models (LLM)
have taken science and society by storm in the recent past 
\cite{Vaswani,Jumper,Lecun}.
Much of this explosive growth is driven by the 
``no-wall'' finding, namely
the fact that the learning capability of LLM-based chatbots 
keeps increasing with size, whence the mantra 
``bigger is better'' which has driven much
of the current leading-edge LLMs research.

Let us recapitulate the main ideas behind the ``no-wall'' finding. 

The scaling law of LLMs is usually expressed in the form

\begin{equation}
\label{NOWALL}
L(N) = A/ N^{\alpha}\:,
\end{equation}

where $L$ is the loss function, a measure of the departure of the
LLM outcome from the desired target, and $N$ stands for 
data ($D$), number of parameters ($P$) and computational cost ($C$),
respectively, each characterized by its constant 
prefactor $A$ and exponent $\alpha$ \cite{Kaplan,Bahri}.
A positive value of the exponent indicates that the loss function
decreases for increasing data size, so that using larger data sets 
leads to a closer match to the desired target.
Formally, the so-called ``wall'' corresponds to the opposite 
regime, in which more data lead to a larger discrepancy between
model and target, as formally reflected by a negative exponent.
The LLMs exponents so far have remained positive across several 
decades, whence the enthusiastic claim that ``there is no wall,''
support a strategy oriented towards ever larger (and energy consuming) 
LLM applications \cite{Nowall}.
 
While the accomplishment is remarkable under all 
scientific counts, many researchers
have observed that the LLMs scaling exponents are 
very small, typically in
the range $0.05 \div 0.10$, pointing to a regime 
of ``diminishing returns'' for pre-trained LLMs.
In a recent paper, two of the present authors have pointed 
out what ``diminishing returns'' really means in actual practice: an exponent $0.1$ means
that cutting the loss function down by a factor 2 requires $2^{10} = 1024$
more resources \cite{CSLLM}.
It was then argued that, while formally wall-free, such regime is simply
unsustainable, whence the need of new directions paying more attention
to physical insight and compliance with world models than muscular
leverage of the number of parameters \cite{CDH,SC19,SH,VAFA}.

Among the critical feedback spawned by this simple  
observation, a recurrent one is that the 
loss function cannot be paralleled to numerical discretization 
errors, the reason being that the ML procedures do not necessarily 
aim at sending the loss function to zero, but are typically stopped 
before that limit is approached, usually in order to 
forestall overfitting and ensuing problems in generalizing, i.e.,
the ability to reproduce unseen targets \cite{Lecun}.

In the following, we argue that such criticism does not change
the conclusion on the unsustainability of LLMs scaling exponents.

\section{The loss function as a pseudo-metric of accuracy}

The approximation error associated with a numerical method employing
$N$ degrees of freedom, say the grid discretization of a PDE,
usually follows an asymptotic scaling relation of the form
\begin{equation}
E(N) = A/N^a\;,
\end{equation}
where $a>0$ is the order of accuracy and $A$ a prefactor
measuring the ``critical'' size $N_c = A^{1/a}$ above which 
the error starts to show a power-law decay. 
Most grid methods work around $a=2$, while stochastic particle 
methods, such as Monte Carlo, feature $a \sim 1/2$.
Note that $a=2$ means that reducing the error by a factor $2$ takes
just $\sqrt{2}$ more resources ($N$), while with $a=1/2$ this number
is $4$, which is generally regarded as poor convergence. 
 
The above relation encodes a basic requirement on any well-posed
numerical scheme, called "Consistency", namely that upon 
sending $N \to \infty$ the error should
vanish, so as to reproduce the original target, generally the 
analytical solution of a continuum differential equation.
Any non zero value $E_{\infty} \equiv E(N \to \infty)$ (the "pedestal"), 
or, worst, increasing error at increasing resolution (the "wall")
is regarded as an anomaly, usually caused by some form of 
ill-posedeness of the numerical discretization, typically the breaking
of a basic continuum symmetry by the discrete scheme.

The above relation bears a direct analogy to the LLM scaling law
\cite{Kaplan}, whence the simple conclusion that an exponent
$\alpha=0.1$ is simply unsustainable.
Claiming that this analogy is flawed because machine learning practices
do not concern themselves with the limit $L \to 0$ as
$N \to \infty$ is tantamount to saying that consistency, a prime 
requirement for any well-posed numerical method, is not relevant to 
machine learning as a scientific discipline.

While it is not hard to see reasons why one might be happy to settle with
``small enough'' values of the loss function (the so-called
early-stop empirical practice), it remains undeniable
that dismissing consistency in favor of ambiguous 
``small enough'' criteria, leaves much to be desired in terms of
reliability of the methodology as a systematic scientific method. 
This is especially true with regard to the ability to generalize
to unseen data, which is the essence of true learning as
opposed to mere memorization of the training data. 
Yet, it is true that matching a given set of data is not equivalent
to reproducing the solution of a continuum PDE, so let us 
accept the loss function as a sort of empirical {\it pseudo}-metric 
in no need to comply with the consistency requirement.
In the following we shall argue that even this ``lenient perspective'' 
does not affect the claim of unsustainability made in \cite{CSLLM}.
The point is no longer consistency, but efficiency, namely
how fast does the error decrease at increasing resources, which is precisely
dictated by the actual value of the scaling exponent.

\section{The pedestal effect}

Recently, a group of authors \cite{GORE} have built on the 
so called Chinchilla scaling \cite{Hoffmann} 
to argue that LLM scaling exponents published by
Anthropic and subsequent works are ``biased'' by the fact of ignoring a
``pedestal'' in the scaling laws, namely the fact that the loss 
function does not vanish in the "continuum limit" $N \to \infty$.

To appreciate the point, let us begin by casting the loss function 
in the form
\begin{equation}
L(x) = L_0 + (L_1 -L_0) \;x^{\alpha}\;,
\end{equation} 
where we have set $x \equiv 1/N$ for the sake of convenience, so 
that $L_0=L(x=0)$ is the continuum limit and 
$L_1 = L(x=1)$ is the large-scale limit $N=1$.
Note that $\alpha>0$ denotes the  ``no-wall'' regime in which $L$ grows with $x=1/N$.

Clearly, the pedestal $L_0$  introduces its own
exponent, $\alpha=0$, so that the ``effective'' 
exponent associated with the above relation must necessarily lie
between $0$ and $\alpha$, the precise form of the transition
depending on the ratio $p=L_0/L_1$.
To highlight the point, let us rescale $L \to L/L_1$ and write
\begin{equation}
\label{CHIN}
L(x;p) = p \; x^0 + (1-p) \;x^{\alpha}\;.
\end{equation} 

Next, let us define a ``running'' scaling exponent 
associated to a given pedestal $p$ as
\begin{equation}
\alpha_p(x) =  \frac{x L'(x,p)}{L(x,p)}\;,
\end{equation}
where prime denotes derivative with respect to  $x$.
Clearly, the running exponent $\alpha_p(x)$ returns a constant 
only in the case of a single-exponent power law behavior,
specifically $\alpha_{p=0}(x)=\alpha$ and $\alpha_{p=1}=0$.

Such function is reported in Fig.\ 1, for $p=0.01,0.05,0.1,0.2$
for the case of Chinchilla scaling $\alpha \equiv \alpha_C \sim 1/3$.
This figure shows that the transition from the 
Chinchilla (C) to the Anthropic (A) with $\alpha \sim 0.05 \div 0.1$
regime is rather sharp, indicating that it does not take extremely 
large datasets to transit from ``high'' (C) to ``low'' (A) scaling regimes. 
The above observation highlights the major relevance
of the pedestal effect in assessing the scaling performance of LLMs.

\begin{figure}
\centering
\includegraphics[scale=0.3]{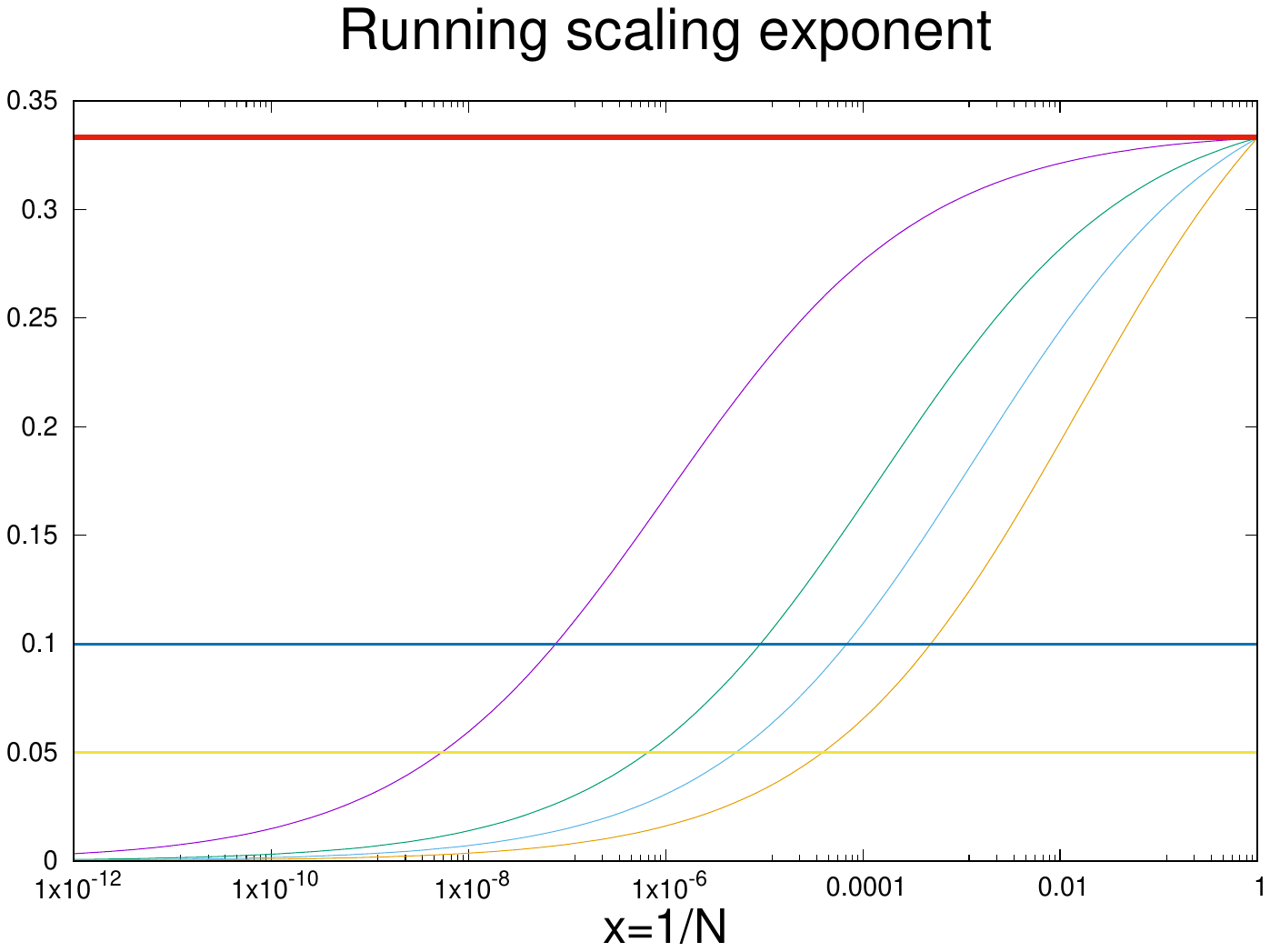}
\caption{The running scaling exponent as a function of $1/N$ for 
$\alpha_C=1/3$ (solid horizontal line) 
and $p=0.01,0.05,0.1,0.2$ (right to left to top).
The figure clearly shows that transition from $1/3$ 
to the A-region $0.05 < \alpha < 0.1$ takes place for 
sizes well below the largest LLM applications. 
Whence the dominance of the pedestal effect in controlling 
the scaling properties of such applications.
The values are divided by $(1-p)$ to ensure the condition
$\alpha(x=1,p)=1$ for any value of $p$.
}
\end{figure}

The same idea can be formulated by introducing
a critical threshold $x_{crit}(p)$ below which
the low-exponent A regime becomes the dominant one.
Based on the expression (\ref{CHIN}), this occurs under 
the condition 
$$
p \gg (1-p)x^{\alpha_C}\;,
$$ 
namely, whenever 
\begin{equation}
x \ll x_{crit}(p) = (\frac{p}{1-p})^{1/\alpha_C}\;.
\end{equation}
Hence, for $\alpha_C = 1/3$, and switching back
to the size $N$, we obtain 
\begin{equation}
\label{NCRIT}
N \gg N_{crit}(p) = (\frac{1-p}{p})^{3}\;. 
\end{equation}
Fig.\ 2 shows the boundary $N_{crit}(p)$ between
the C and A regimes as a function of $p$. 
From this figure, it is clear that it takes very small 
values of the pedestal in order for the C regime to be 
the dominant for large data sets. 
Differently restated, by inverting the relation (\ref{NCRIT}), we obtain
\begin{equation}
\label{NCRIT}
p \ll p_{crit}(N) = \frac{1}{1+N^{1/3}} \sim N^{-1/3}\;.
\end{equation}
Hence, even for a moderate size dataset with 
$N=10^6$, we already have $p_{crit} \sim 10^{-2}$. 
Since the loss function in most LLMs applications decays by about half a decade 
over several decades in $N$, the pedestal value 
is well above $0.01$, showing that the Chinchilla exponent
is not relevant to the scaling performance 
of large datasets such as the ones used in modern LLMs applications. 

\begin{figure}
\centering
\includegraphics[scale=0.3]{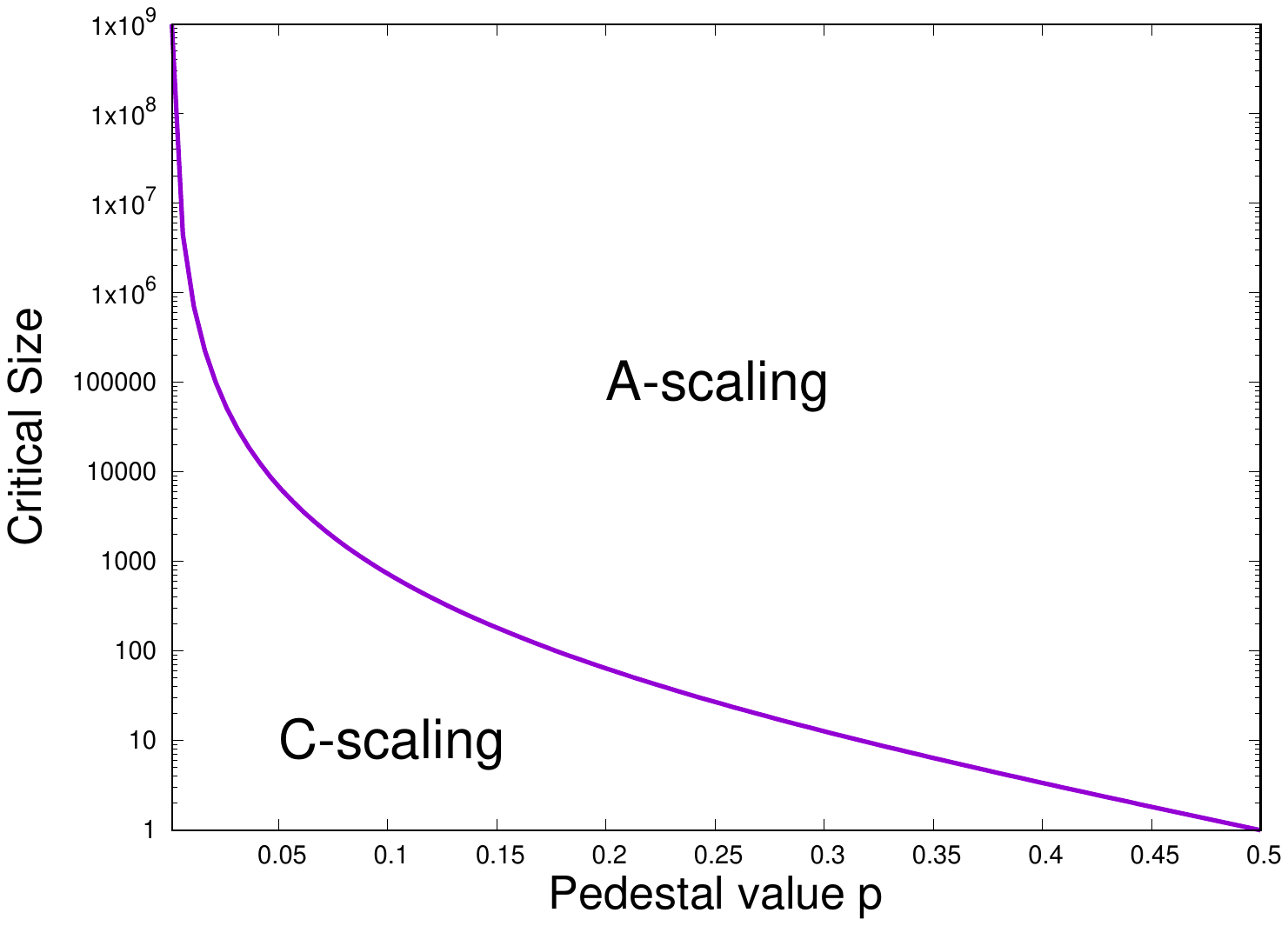}
\caption{The critical boundary $N_{crit}(p)=(\frac{1-p}{p})^3$ between 
C and A scaling regions, as a function of the pedestal $p$.
The C-scaling is only relevant below the critical boundary, hence 
for relatively small datasets size, unless the pedestal is 
made unrealistically small.
}
\end{figure}
  
\section{Theoretical prediction of the scaling exponents}

Having pinpointed the importance of the pedestal in assessing 
the effective scaling properties of LLMs, the natural question
is to look for a theoretical explanation of their value.
A simple yet quite convincing explanation was provided 
by Sharma and  Kaplan (SK hereafter) \cite{SK}, whose work
lends further credit to the relevance of the 
A-exponents as a proper measure of LLMs scaling performance.

In this paper present a simple and convincing toy-theory, supported
by experimental data, that the LLM scaling 
exponents obey the following lower bound,
\begin{equation}
\label{SK}
\alpha \sim \frac{4}{d}\;,
\end{equation} 
where $d$ is the Intrinsic Dimension of the manifold, 
namely the number of independent coordinates required to describe the 
manifold where most data reside \cite{Ansuini}.

Note that for many complex applications $d \ll  D$, $D$ being the 
dimension of the embedding space.
The above relation is remarkable for its simplicity, elegance 
and robustness across a broad ensemble of numerical datasets.

A few comments are in order.

{\it First,\/} note that the SK relation refers to 
concrete LLMs practice, hence it is fully consistent with 
the pseudo-metric perspective discussed above.

{\it Second,\/} let us recall that in $D$ dimensions, a grid 
discretization method of order $a$, features an exponent
\begin{equation}
\label{COD}
\alpha(D) = \frac{a}{D}\;.
\end{equation}
This results by the sheer observation that the volume $V(D)$
of a D-dimensional region of space of diameter $\delta$ scales like 
$V(D) \sim \delta^D$, and standard (non-adaptive) 
grid discretization treats all portions of the D-dimensional
space on the same footing, regardless of whether or not they
host any interesting information process (typically they do not).
This is the infamous Curse of Dimensionality (CoD) \cite{Bellmann}.
 
{\it Third,\/} the expression (\ref{COD}) invites a natural analogy with 
the SK relation with  $D=d$ and $a=4$.
This means that LLMs can be paralleled to adaptive grid refinement
methods with $a=4$ in a $d$-dimensional feature space. 
By adaptive, we mean methods which place the
numerical degrees of freedom "on demand", i.e.,
there where the relevant information is located
and not everywhere in a D-dimensional region of space.
This ability to spot the relevant low-dimensional manifold  
where the computational resources should be focused is essential
to (partially) tame the CoD, possibly one of the major achievements 
of LLMs research. 

Let us expand on the above by revisiting the main ideas behind 
the SK formula and connect them with the phenomenology of fluid turbulence.

\section{The SK model and connections with turbulent fractals}

Let us consider a $d$-dimensional cube of side $1$ and fill it in with
$N(s)$ cublets of side $s$. 
By definition $N(s)s^d = 1^d=1$, whence
\begin{equation}
\label{NS}
N(s) = s^{-d}\;.
\end{equation}
Next, let us define the loss function as the numerical error associated
with a piecewise representation $f_N(x)$ of a {\it smooth\/} function 
$f(x)$ within the unit side cube
\begin{equation}
L(s) =  \int |f(x)-f_N(x)|^2 d_dx
\end{equation}
Since the function $f(x)$ is smooth, there is always 
a position $x'$ in the hypercube, such that $f_N(x)=f(x')$.
Hence, by virtue of smoothness, the integrand in the above
integral is bounded by $gsd^{1/2}$, where $g$ is a constant
proportional to the gradient of $f$ and $sd^{1/2}$ is the 
diameter of the hypercube.
As a result, we obtain 
$L(s) \sim g^2 s^2 d$, namely, based on (\ref{NS}),
\begin{equation}
\label{LS}
L(N) \sim N^{-2/d}\;,
\end{equation}
whence $\alpha=2/d$.
The same argument as applied to a piecewise linear representation
of $f(x)$ returns $s^4$ instead of $s^2$, whence the
sought $4/d$ exponent.
Essentially the scaling exponent is dictated by the smoothness (roughness)
of the target function, the accuracy of the discretization
and the dimensionality of the intrinsic manifold.

KS justly evoke a connection between adaptive refinement and fractal sets, and 
in the following we expand such analogy in semi-quantitative terms building on
standard arguments from the fractal theory of fluid turbulence \cite{FRISCH}.
The purpose is by no means to claim a direct connection between LLMs and
turbulence, but just to emphasize that many/most complex phenomena do
not give rise to smooth signals \cite{SS22} and, accordingly to (\ref{LS}),
this impacts directly into the value of the corresponding scaling exponents. 
More precisely, rough signals lead to smaller scaling exponents 
than smooth ones.

\subsection{Scaling exponents of three-dimensional homogeneous incompressible turbulence}

Let us consider a turbulent eddy (active degree of freedom  turbulent
flow) of size $l$. More precisely, let $v(l)$ be the increment of the
velocity across a turbulent eddy of size $l$, i.e.,
$v(l) = |v(x+l)-v(x)|$ where $x$ runs over the geometrical
region occupied by the fluid (for a homogeneous fluid the dependence
on $x$ drops out).  
Based on Kolmogorov's 1941 theory \cite{K41}, the energy supplied to the
flow at large scales is transferred to small scales, virtually with 
zero dissipation, through the process of energy cascade. 
Large eddies break-up into smaller daughter 
eddies, which further break-up into even smaller grand-daughter eddies 
and so on down the line until eddies are sufficiently small 
for dissipation to take over.

According to K41, the energy flux is the same across scales, 
hence the energy dissipation rate writes as
\begin{equation}
\label{K41} 
\epsilon(l) = v^3(l)/l = const.\;, 
\end{equation}
where we have taken the lifetime of the above eddy 
as $\tau(l) \sim l/v(l)$.

This implies $v(l) \sim l^{1/3}$, i.e., the turbulent flow
is {\it not} smooth, but exhibits a scaling (Hurst) exponent 
$h=1/3$, as opposed to a smooth (differentiable) signal 
featuring $h=1$.
This alone brings the scaling exponent down by a factor 
$h$, i.e., $\alpha \sim -4h/d$, with no need of invoking
fractal structures, just the fact that the signal is not differentiable. 

The connection to fractals arises by the observation that 
K41 theory captures many features of turbulence, but fails to describe
intermittency, namely the occurrence, here and there, of abrupt
bursts of activity. To account for this feature, one postulates that
energy dissipation does not act as a space-filling process, but rather
occurs on a set of fractal dimension $d<D$ where $D$ is the dimension of
the embedding space, for fluids typically $D=2$ or $D=3$.

To compute the fractal dimension $d$, an argument pretty 
]similar to KS is usually adopted \cite{FSL}.
Namely, one introduces the probability $\beta(l)$ that a small
sphere (or cublet) of radius $l$ intercepts a $d-dimensional$ object within
a three-dimensional ($D=3$) cube of side $1$.
Such probability is readily shown to scale as
$$
\beta(l) \sim l^{D-d}\;,
$$
where $D-d$ is the so called co-dimension.
Next, one assumes that each time a turbulent eddy of size $l$ breaks
down in a series of smaller eddies, only a fraction $\beta(l)$  
remains active, namely amenable to further breakup. 
Hence, the kinetic energy available to sustain the energy
cascade at scale $l$ is $E(l) = \beta(l) v^2(l)$.
 
Combining this with (\ref{K41}), one obtains
\begin{equation}
v(l) \sim \epsilon_0 l^{1/3} \beta(l)^{-1/3}\;,
\end{equation}
yielding an effective exponent
\begin{equation}
\label{HBETA}
h = 1/3 - (D-d)/3 = \frac{d-2}{3}\;,
\end{equation}
where we have used $D=3$.
This shows that the K41 exponent $h=1/3$ corresponds to
a space-filling dissipation process with $d=D=3$, whereas intermittency 
occurs on a fractal set of dimension $d<D=3$ so that $h<1/3$.

Differently restated, the turbulent energy is a space-filling process
and in the K41 picture so is energy dissipation: dissipation 
``does not hide,'' the intrinsic manifold fills up the entire embedding space. 
Even so, the velocity signal is pretty rough, with a scaling exponent $h=1/3$.

In the case of intermittency, dissipation is no longer space filling, and
the velocity signal gets further roughness in proportion to the co-dimension
of the fractal set where dissipation takes place, according to the
expression (\ref{HBETA}).

In homogeneous incompressible fluid turbulence one finds 
$d \sim 2.8$ (the so-called beta model discussed above) 
corresponding to $h \sim 0.27$, but other models deliver different 
values, all above $2$, since  $d<2$ would imply $h<0$, denoting
a singularity in the velocity field, something that cannot 
happen in a homogeneous incompressible flows.

In passing, it is worth mentioning that turbulence is in fact 
multifractal, meaning by this that dissipation inhabits a 
whole set of different manifolds, each with its own scaling 
exponent \cite{FRISCH,BRISCO}.

The implications for LLMs are as follows.

For non-smooth processes, such as turbulence and many other
complex dynamical systems with $h<1$, the scaling 
exponents become even smaller, namely
\begin{equation}
\alpha \sim \frac{4h}{d}\;.
\end{equation}
Differently restated, the intrinsic dimensions of non-smooth
processes are magnified by a factor $1/h$, which spells
even more trouble for LLM performance for non-smooth data.
  
On the other hand, LLM appear to be amazingly efficient 
in compressing  the relevant information to 
low-dimensional manifolds. 
While turbulence only  ``needs'' to go from $D=3$ to $d>2$ in 
order to pin-down
the dissipative manifold, LLMs accomplish far larger 
dimensional compression
tasks: the embedding dimension of feature space in modern 
chatGPT's scores in the 
several thousands, as opposed to an intrinsic dimensions $d \sim 100$,
corresponding to a quite impressive dimensional compression 
$d/D \sim O(10^{-2})$. 

The ability of LLMs to pin down and represent information 
under such extreme dimensional compression regimes is quite remarkable.
      
\section{Conclusion}

If the SK analysis is anything to go by, and we see no reason 
to doubt it, we must conclude that any dataset with 
$d>40$ is bound to feature $\alpha < 0.1$, hence doomed to unsustainability,
regardless of whether or not the loss function should 
be paralleled to a numerical discretization error.
As shown in this note, the situation is only going
to get worse in the case of non-smooth data, as it 
is often the case for complex systems.
This is the actual state of affairs: notwithstanding the absolutely 
amazing ability of LLMs to chase the information hidden
in extremely tiny corners of ultra-dimensional feature spaces, the
dimension of the intrinsic manifolds is still too high to
be sustainable.  
This is why going down the ``bigger is better'' avenue under the drive of the
``no-wall'' excitement is a sure recipe for energy burnout.
Quite possibly, a return to foundational physics-aware AI models, the so
called "world models" \cite{VAFA} appears to offer a 
way more promising strategy.  

This work was partly supported by the Research Council of Norway through its Centers of Excellence funding scheme, project number 262644. AH furthermore acknowledges funding from the European Research Council (Grant Agreement 101141323 AGIPORE).









%

\end{document}